# Contextual Embedding Evidence for Main–Light Verb Distinctions in Urdu

**Farah Adeeba**
Cluster of Excellence "The Politics of Inequality"
University of Konstanz
farah.adeeba@uni-konstanz.de

**Miriam Butt**
Department of Linguistics
University of Konstanz
miriam.butt@uni-konstanz.de

## Abstract

Urdu light verbs contribute schematic event-structural meaning while remaining lexically related to corresponding main verbs. This study tests representational predictions derived from Butt's analysis using contextual embeddings from UrduBERT, DunbaaBERT, and multilingual BERT across 1,126 naturally occurring sentences containing seven Urdu verbs. Main and light uses show significant representational separation in all 21 verb–model comparisons. At the same time, same-lemma main and light centroids are consistently closer than mismatched main–light lemma pairs, supporting continued lexical relatedness. In a seven-way prediction task restricted to light uses, verb identity remains recoverable after the target is masked, with UrduBERT achieving 0.866 accuracy and 0.852 macro-F1. UrduBERT also retains 0.782 accuracy under a preceding-form-disjoint evaluation, indicating generalization beyond repeated local verb combinations. These findings provide computational evidence consistent with Butt's account that Urdu light verbs differ systematically from their main uses while retaining lemma-specific and verb-specific representational structure.

## 1 Introduction

Light verb constructions are a central feature of Urdu grammar. In these constructions, a lexical verb (V1) combines with a second verb (V2) that contributes aspectual, completive, attitudinal, or event-structural meaning while exhibiting reduced lexical content relative to its corresponding main-verb use. For example, consider the two uses of *gayā* 'went' in (1)–(2). In (1), *gayā* functions as a main verb and expresses literal motion. In (2), it combines with the lexical verb *so* 'sleep' and functions as a light verb, contributing a completive or inchoative interpretation rather than its independent motion meaning.

(1) وہ گھر گیا
*vo ghar gaya*
'He went home.'

(2) بچہ سو گیا
*bacca so gaya*
'The child fell asleep.'

This study is grounded in Butt's analysis of Urdu light verbs (Butt, 1995, 2003, 2010) and in the lexical-relatedness account developed by Butt and Lahiri (2013). Although this contrast has received extensive linguistic analysis, it remains unclear whether contextual language models encode the distinction and whether their representational geometry reflects predictions derived from these accounts.

This study examines 1,126 naturally occurring Urdu sentences containing seven canonical light verbs. Representative main and light uses of all seven verbs examined in this study are provided in Appendix E. Three encoder-based models are compared: UrduBERT, DunbaaBERT (Maab et al., 2026), and multilingual BERT (Devlin et al., 2019). The central question is: how far apart are the main and light uses of the same Urdu verbs in contextual embedding space? If the two uses are represented similarly, their embeddings should overlap substantially. If the models encode the distinction, the two usage types should show measurable separation or occupy distinguishable regions of the vector space.

We investigate this question through complementary geometric, supervised, and unsupervised analyses. Centroid cosine distance measures the magnitude of separation, while silhouette scores, logistic probing, KMeans clustering, and PCA characterize the structure and recoverability of the distinction. Two supplementary analyses examine whether the observed geometry is compatible with further predictions derived from Butt's account: same-lemma main and light centroids are compared with mismatched cross-lemma pairs, and light-verb identity is predicted after masking the target form, including under a preceding-form-disjoint split. Embedding distance is treated as a measure of representational differentiation rather than as a direct measure of semantic distance.

The contributions are threefold. First, we provide a multi-model analysis showing that the main–light distinction is systematically reflected in contextual embedding space across seven Urdu verbs. Second, we show that same-lemma main and light centroids remain closer than mismatched main–light lemma pairs across all three models, a pattern compatible with continued lexical relatedness. Third, we show that individual light verbs retain distinguishable contextual profiles after target masking and, particularly for UrduBERT, under a preceding-form-disjoint evaluation. Together, these findings provide computational evidence consistent with specific representational consequences of Butt's analysis, without directly establishing a particular formal lexical-entry structure or diachronic pathway.

## 2 Background and Related Work

The theoretical foundation for the analysis of Urdu light verbs was established by Butt (1995). On this account, light verbs modify event structure and contribute weak but non-vacuous schematic content rather than the full lexical meaning associated with their main uses. Despite this semantic reduction, light verbs remain within the lexical domain and retain verb-specific contributions. This position was subsequently developed in further work (Butt, 2003, 2010).

Butt and Lahiri (2013) add a diachronic dimension, arguing that light verbs are pertinacious and that main and light uses remain related through a shared lexical source rather than following a simple progression toward semantically empty grammatical markers. This analysis motivates two representational expectations relevant to the present study. Main and light uses should be systematically differentiated because their contextual contributions differ, but same-lemma uses may nevertheless retain greater proximity than combinations involving different verb lemmas. Moreover, if individual light verbs preserve schematic event-structural content, they should not collapse into a homogeneous and interchangeable class.

Corpus evidence further establishes the grammatical and distributional importance of the construction. Ahmed (2010) shows that a relatively small inventory of verbs accounts for a substantial proportion of naturally occurring Urdu light-verb constructions. The seven verbs examined here are drawn from this theoretically and empirically motivated inventory. Computational approaches to identifying Urdu complex predicates have also been explored using bigram extraction over corpus data, establishing that surface co-occurrence patterns can partially recover the V+V constructions that form the basis of light verb annotation (Butt et al., 2012)

Contextual language models provide a way to examine whether different uses of the same surface form receive systematically different representations. A recent study of English introduced a minimal-pair dataset for testing whether language models distinguish light-verb and full-verb uses and reported separable representational patterns (Franzon et al., 2026). Related probing work on English multi-word verbs has likewise shown that contextual representations can distinguish verb-particle construction types, while classifier accuracy and geometric separability need not align (Kissane et al., 2025). Together, these studies demonstrate the feasibility of probing verbal-construction distinctions computationally, but they focus on English and do not evaluate the additional lexical-relatedness and verb-specificity predictions motivated by the Urdu literature. For Urdu specifically, (Hautli-Janisz et al., 2025) evaluate GPT as an automatic annotator of light verb constructions, demonstrating that the main–light distinction is behaviorally accessible to generative models. The present study is complementary in focus: rather than annotation behavior, the question examined here is whether the distinction is encoded in the representational geometry of encoder-based models, and whether that geometry reflects further predictions derived from Butt's account.

## 3 Dataset

We constructed our dataset from an in-house Urdu corpus comprising approximately 1.6 GB of raw text, compiled from multiple sources including the OSCAR 2019 corpus,[1] GitHub, and Kaggle repositories, with deduplication applied to ensure data quality. The evaluation sentences were drawn from a held-out portion that was not used during UrduBERT pre-training. We selected seven high-frequency Urdu verbs known to exhibit both main and light verb behavior in the linguistic literature: *āyā* (آیا, 'came'), *uṭhā* (اٹھا, 'rose/got up'), *baiṭhā* (بیٹھا, 'sat'), *paṛā* (پڑا, 'fell/lay'), *diyā* (دیا, 'gave'), *gayā* (گیا, 'went'), and *liyā* (لیا, 'took'). These verbs were selected from the inventory of Urdu light verbs identified by Butt (1995) and further studied by Ahmed (2010), who analysed their distributional frequencies in a corpus of over 14 million tokens of naturally occurring Urdu text. Sentences were extracted from the corpus where the target verb appeared in sentence-final position, consistent with Urdu's SOV word order. Annotation followed a strict three-way agreement protocol involving GPT-5.1 and two expert human annotators, both authors of this paper with expertise in Urdu linguistics. In the first stage, each sentence was classified by GPT-5.1 using a prompt grounded directly in Butt's 1995 linguistic criteria. The model assigned one of three labels: *main*, where the word immediately preceding the target verb is a noun or noun phrase and the verb expresses its full lexical meaning (e.g. *us ne mujhe qalam diyā*, 'he gave me a pen', where *diyā* retains its full meaning of transfer); *light*, where the immediately preceding element is another verb stem forming a V1+V2 compound and the target verb contributes only aspectual or completive meaning (e.g. *baccā so gayā*, بچہ سو گیا, 'the child fell asleep', where *gayā* adds completion rather than motion); and *skip*, reserved for noun-verb complex predicates such as *kām karnā* (کام کرنا, 'to work') or *faislā karnā* (فیصلہ کرنا, 'to decide'), as well as structurally ambiguous cases. This three-way distinction is directly motivated by Butt's 1995 characterization of Urdu light verbs, ensuring that complex predicates and ambiguous instances did not contaminate the final categories. In the second and third stages, the two human annotators independently reviewed each GPT-assigned label, recording agreement or disagreement. Only sentences for which all three sources, namely GPT-5.1, Annotator 1, and Annotator 2, were in full agreement were retained; any sentence attracting even a single disagreement was discarded. The final dataset comprises **1,126** sentences distributed across the seven target verbs, as shown in Table 1.

| Verb | Light | Main | Total |
|---|---|---|---|
| آیا (*āyā*) | 58 | 88 | 146 |
| اٹھا (*uṭhā*) | 84 | 28 | 112 |
| بیٹھا (*baiṭhā*) | 59 | 28 | 87 |
| پڑا (*paṛā*) | 158 | 99 | 257 |
| دیا (*diyā*) | 131 | 52 | 183 |
| گیا (*gayā*) | 68 | 113 | 181 |
| لیا (*liyā*) | 106 | 54 | 160 |
| **Total** | **664** | **462** | **1126** |

Table 1: Dataset statistics showing light and main verb instance counts for each of the seven target verbs.

## 4 Methodology

The methodology evaluates three representational predictions. The first concerns differentiation between main and light uses of the same verb. The second concerns continued lexical relatedness between those uses. The third concerns verb-specific structure within the light-verb inventory. The tense auxiliary comparison is retained as an exploratory grammatical reference condition and is not treated as a direct scale of semantic distance. All experiments were conducted on a single NVIDIA H100 GPU with a 40 GB memory partition, with models loaded sequentially in evaluation mode.

### 4.1 Models

Three encoder-based models are compared, selected to represent a spectrum from Urdu-specific to multilingual pre-training.

**UrduBERT** is a BERT-base model trained from scratch on a 5.8 GB deduplicated Urdu corpus (OSCAR, GitHub, and Kaggle) using masked language modeling for 20 epochs. A WordPiece tokenizer with a 64k-token vocabulary was trained on the same data. UrduBERT[2] was developed as part of the present study; no prior published reference exists for this model, and full pre-training details are provided in Appendix D.

[1] OSCAR (Open Super-large Crawled ALMAnaCH coRpus) is a multilingual corpus derived from Common Crawl via language classification and filtering; https://huggingface.co/datasets/oscar-corpus/oscar

[2] Model is available at https://huggingface.co/farahadeeba/urdu-bert-64k

**DunbaaBERT** (Maab et al., 2026) is a RoBERTa-style (Zhuang et al., 2021) model trained on a 17 GB Urdu corpus comprising mC4, OSCAR, Wikipedia, and NLLB data, using whole-word masking and a 52k Byte-BPE vocabulary. We use the 52k-vocabulary variant.

**Multilingual BERT** (mBERT; Devlin et al. 2019) serves as a multilingual comparison model. It was pre-trained on Wikipedia text from 104 languages and was not specifically optimized for Urdu.

### 4.2 Embedding Extraction

For each sentence, the contextual embedding of the target verb is extracted from the final hidden layer. Verb token positions are identified through the tokenizer's offset mapping by matching token character spans against the final occurrence of the target verb in the original sentence. Where a target spans multiple subword tokens, attention-weighted pooling is applied: each subword is weighted by the mean attention it receives across all heads in the final layer, normalized over the target positions. A comparison with mean pooling produced near-identical results across models and metrics, with a maximum probe-F1 difference of 0.003.

### 4.3 Main–Light Representational Separation

For each verb and model, representational separation is quantified as the cosine distance between the main-use and light-use embedding centroids. This measure captures distributional divergence in embedding space and is not interpreted as a direct estimate of semantic distance. A silhouette score using cosine distance is additionally computed to capture both within-class compactness and between-class separation. Statistical significance is assessed with a one-sided label-permutation test using 1,000 permutations under the null hypothesis that main and light labels are exchangeable.

PCA is fitted independently to each verb's embedding matrix for visualization. A class-weighted logistic regression probe predicts main versus light labels using five-fold stratified cross-validation. KMeans clustering with $k = 2$ is applied as an annotation-independent analysis, and agreement with the gold labels is measured using Adjusted Rand Index (ARI). KMeans was run with three random seeds (42, 52, and 62); all runs produced identical evaluation scores, indicating stability across initializations. The probe tests linear recoverability, whereas KMeans tests whether the binary distinction aligns with an unsupervised two-cluster partition. PCA is used only for qualitative visualization; all claims of separation are based on the full-dimensional quantitative analyses.

### 4.4 Same-Lemma versus Cross-Lemma Proximity

To evaluate the prediction of continued lexical relatedness, we compare same-lemma main–light distance with mismatched cross-lemma distance. Let $\mathbf{c}_v^M$ and $\mathbf{c}_v^L$ denote the main-use and light-use centroids of verb $v$. The same-lemma distance is

$$d_{\text{same}}(v) = 1 - \cos(\mathbf{c}_v^M, \mathbf{c}_v^L). \tag{1}$$

For each verb, the cross-lemma reference is the mean of symmetric mismatched distances involving all other lemmas $u \neq v$:

$$d_{\text{cross}}(v) = \frac{1}{2(|V|-1)} \sum_{u \neq v} \left[ d(\mathbf{c}_v^M, \mathbf{c}_u^L) + d(\mathbf{c}_u^M, \mathbf{c}_v^L) \right]. \tag{2}$$

We report $d_{\text{cross}} - d_{\text{same}}$, the rank of the matching light centroid for each main centroid, top-1 matching accuracy, and mean reciprocal rank. An exact one-sided sign-flip test across the seven verbs evaluates whether the difference is systematically positive. We also perform a sentence-level stratified bootstrap with 5,000 resamples, sampling main and light instances independently within each verb, and report percentile 95% confidence intervals for the mean difference. Because the original target form is present in both same-lemma conditions, shared orthographic and lexical identity may contribute to this effect; the analysis is therefore interpreted as compatible with lexical relatedness rather than as a direct test of lexical-entry structure.

### 4.5 Masked Light-Verb Identity Prediction

To evaluate whether individual light verbs retain distinct contextual profiles, we restrict the data to the 664 light-use sentences and train a seven-way class-weighted logistic regression classifier to predict light-verb identity. Two conditions are compared. In the *original-target* condition, the classifier receives the contextual embedding of the overt target verb. In the more restrictive *masked-target* condition, the final occurrence of the target verb is replaced with the model's mask token and the final-layer hidden state at the mask position is used as the representation. The original-target condition is a sanity check for preservation of lexical identity; the masked-target condition tests whether the surrounding context contains verb-specific information after removal of the target form.

Performance is measured using accuracy and macro-F1 under five-fold stratified cross-validation. Results are compared with a majority-class baseline and a uniform seven-class chance level of $1/7$. Significance for the masked condition is assessed with 1,000 label permutations. For each permutation the classifier is refitted using the same fold assignments, and corrected one-sided $p$-values are computed as $(b+1)/(N+1)$, where $b$ is the number of permuted scores at least as large as the observed score and $N = 1000$.

### 4.6 Preceding-Form-Disjoint Evaluation

A grouped evaluation tests whether masked-target prediction generalizes beyond repeated immediate V1–V2 combinations. Each light-use sentence is grouped by the surface form immediately preceding the masked target, yielding 153 unique groups. Five-fold StratifiedGroupKFold cross-validation ensures that no preceding surface form occurs in both training and test data. This analysis is termed *preceding-form-disjoint* rather than strict leave-V1-lemma-out because the groups are automatically derived surface forms and are not fully normalized morphological lemmas. Accuracy and macro-F1 are reported using pooled out-of-fold predictions.

### 4.7 Auxiliary Reference Analysis

As a grammatical control, we randomly sampled 300 unique tokens tagged as tense auxiliaries (AUXT) by the CLE Urdu POS tagger (Urooj et al., 2014) from the held-out corpus. We manually verified their grammatical function and computed a global auxiliary centroid from the retained instances. For each target verb, we calculate $d_{\text{ML}}$ (main–light), $d_{\text{MA}}$ (main–auxiliary), and $d_{\text{LA}}$ (light–auxiliary), together with $r_v = d_{\text{LA}}/d_{\text{MA}}$. A ratio below 1 indicates that the light centroid is closer to the auxiliary reference than the main centroid is. Because these categories differ in lexical identity and grammatical function, the analysis is exploratory and is not treated as a direct measure of semantic distance.

## 5 Results

### 5.1 Main–Light Representational Separation

Table 2 reports the cosine distance and silhouette score between main-use and light-use centroids for each verb. All 21 cosine distances are significant at $p < 0.001$ under 1,000-permutation label tests.

| | UrduBERT | | DunbaaBERT | | mBERT | |
|---|---|---|---|---|---|---|
| **Verb** | Dist | Sil | Dist | Sil | Dist | Sil |
| āyā | 0.139 | 0.153 | 0.024 | 0.074 | 0.054 | 0.079 |
| uṭhā | 0.188 | 0.326 | 0.041 | 0.291 | 0.026 | 0.088 |
| baiṭhā | 0.203 | 0.342 | 0.053 | 0.176 | 0.045 | 0.102 |
| paṛā | 0.284 | 0.391 | 0.054 | 0.248 | 0.056 | 0.166 |
| diyā | 0.136 | 0.223 | 0.020 | 0.174 | 0.040 | 0.107 |
| gayā | 0.141 | 0.230 | 0.025 | 0.178 | 0.054 | 0.092 |
| liyā | 0.152 | 0.239 | 0.022 | 0.210 | 0.042 | 0.102 |
| **Mean** | **0.177** | **0.272** | **0.034** | **0.193** | **0.045** | **0.105** |

Table 2: Cosine distance (Dist) and silhouette score (Sil) per verb and model.

UrduBERT produces the largest centroid distances across all seven verbs (mean 0.177), followed by mBERT (0.045) and DunbaaBERT (0.034). UrduBERT also obtains the highest mean silhouette score (0.272). Among verbs, *paṛā* (پڑا) shows the largest separation, whereas *diyā* (دیا) and *āyā* (آیا) show smaller distances. These values establish a gradient in representational separation, not a direct scale of semantic distance. Figures 1 and 2 present the per-verb PCA projections.

### 5.2 Linear Separability and Unsupervised Clustering

Table 3 reports cross-validated probe performance. All models substantially exceed the mean majority-class baseline of 0.675. UrduBERT achieves the highest mean F1 (0.997), followed by DunbaaBERT (0.986) and mBERT (0.952).

| | UrduBERT | | DunbaaBERT | | mBERT | |
|---|---|---|---|---|---|---|
| **Verb** | F1 | Acc | F1 | Acc | F1 | Acc |
| āyā | 1.000 | 1.000 | 0.987 | 0.992 | 0.925 | 0.953 |
| uṭhā | 0.994 | 0.991 | 0.982 | 0.973 | 0.948 | 0.920 |
| baiṭhā | 1.000 | 1.000 | 1.000 | 1.000 | 0.916 | 0.877 |
| paṛā | 0.997 | 0.996 | 0.984 | 0.981 | 0.966 | 0.957 |
| diyā | 0.992 | 0.989 | 0.985 | 0.978 | 0.974 | 0.962 |
| gayā | 0.993 | 0.994 | 0.971 | 0.978 | 0.964 | 0.972 |
| liyā | 1.000 | 1.000 | 0.995 | 0.994 | 0.972 | 0.963 |
| **Mean** | **0.997** | **0.996** | **0.986** | **0.985** | **0.952** | **0.943** |

Table 3: Class-weighted logistic probe results under five-fold stratified cross-validation.

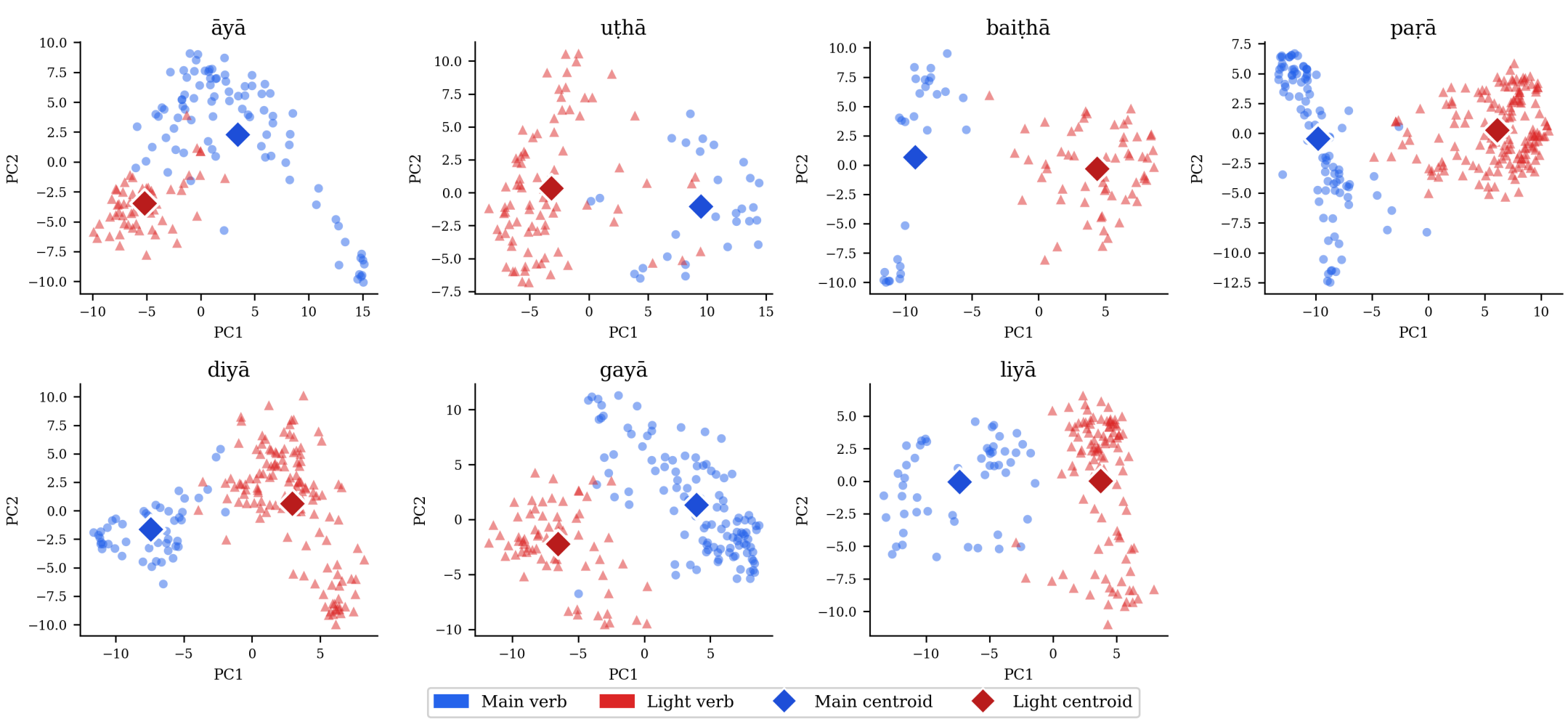

Figure 1: UrduBERT PCA projections for the seven verbs. Main and light uses show clearer visual separation than in the other models; diamonds mark class centroids.

Table 4 shows that unsupervised two-cluster alignment varies substantially by model and verb. UrduBERT reaches a mean ARI of 0.778, while DunbaaBERT and mBERT reach 0.017 and 0.141, respectively. Thus, high linear recoverability does not necessarily imply that the distinction forms two dominant spherical clusters.

| Verb | UrduBERT | DunbaaBERT | mBERT |
|---|---|---|---|
| āyā | 0.359 | −0.019 | 0.018 |
| uṭhā | 0.690 | 0.125 | −0.001 |
| baiṭhā | 0.954 | −0.023 | 0.237 |
| paṛā | 0.893 | 0.085 | 0.382 |
| diyā | 0.890 | 0.010 | 0.043 |
| gayā | 0.658 | −0.002 | 0.039 |
| liyā | 1.000 | −0.053 | 0.270 |
| **Mean** | **0.778** | **0.017** | **0.141** |

Table 4: Per-verb ARI from KMeans clustering with $k = 2$.

## 5.3 Same-Lemma versus Cross-Lemma Proximity

Table 5 reports the comparison between matching and mismatched main–light lemma pairs. For every verb in every model, the same-lemma distance is smaller than the corresponding mean cross-lemma distance. The matching light centroid is the nearest light centroid for all seven main centroids in all three models, yielding top-1 accuracy and mean reciprocal rank of 1.0. An exact one-sided sign-flip test over the seven verbs gives $p = .016$ for each model. The result is compatible with continued lexical relatedness between main and light uses:

| Model | Same | Cross | Δ | 95% CI |
|---|---|---|---|---|
| UrduBERT | 0.177 | 0.513 | 0.336 | [0.326, 0.340] |
| DunbaaBERT | 0.034 | 0.091 | 0.057 | [0.054, 0.060] |
| mBERT | 0.045 | 0.327 | 0.281 | [0.275, 0.284] |

Table 5: Mean same-lemma and cross-lemma main–light centroid distances. Δ is cross minus same. Confidence intervals are sentence-level percentile bootstrap intervals from 5,000 resamples.

the two uses are differentiated but remain closer than mismatched lemma pairs. Because the same orthographic target form is present in the matching condition, lexical identity may contribute to the effect.

## 5.4 Masked Light-Verb Identity

Table 6 reports seven-way light-verb identity prediction over the 664 light-use sentences. With the target present, identity is recovered almost perfectly by all models. More importantly, after replacing the target with the model's mask token, all three models remain substantially above the majority baseline of 0.252 and the uniform chance level of 0.143. UrduBERT reaches 0.866 accuracy and 0.852 macro-F1; DunbaaBERT reaches 0.616/0.585; and mBERT reaches 0.633/0.599.

The preceding-form-disjoint evaluation uses 153 unique groups. UrduBERT shows a reduction from 0.866 to 0.782 accuracy, remaining well above the majority baseline of 0.238 and indicating that its predictions generalize beyond repeated immediate V1–V2 combinations. DunbaaBERT decreases

| Model | Original | | Masked | | Form-disjoint | |
|---|---|---|---|---|---|---|
| | Acc | F1 | Acc | F1 | Acc | F1 |
| UrduBERT | 1.000 | 1.000 | 0.866 | 0.852 | 0.782 | 0.748 |
| DunbaaBERT | 0.988 | 0.989 | 0.616 | 0.585 | 0.520 | 0.483 |
| mBERT | 1.000 | 1.000 | 0.633 | 0.599 | 0.571 | 0.532 |

Table 6: Seven-way light-verb identity prediction. Form-disjoint evaluation excludes preceding-form overlap across folds.

from 0.616 to 0.520, and mBERT decreases from 0.633 to 0.571.

### 5.5 Auxiliary Reference Analysis

Table 7 reports the three-way distances and relative auxiliary-proximity ratio for UrduBERT. Triangle visualizations for all three models are provided in Appendix F.

| Verb | $d_{\mathrm{ML}}$ | $d_{\mathrm{MA}}$ | $d_{\mathrm{LA}}$ | $r_v$ |
|---|---|---|---|---|
| āyā | 0.139 | 0.603 | 0.624 | 1.035 |
| uṭhā | 0.188 | 0.695 | 0.671 | 0.965 |
| baiṭhā | 0.203 | 0.815 | 0.850 | 1.043 |
| paṛā | 0.284 | 0.683 | 0.727 | 1.065 |
| diyā | 0.136 | 0.514 | 0.457 | 0.889 |
| gayā | 0.141 | 0.578 | 0.487 | 0.843 |
| liyā | 0.152 | 0.568 | 0.527 | 0.929 |
| **Mean** | 0.177 | 0.637 | 0.621 | 0.967 |

Table 7: Three-way cosine distances — UrduBERT. $d_{\mathrm{ML}}$ = main–light; $d_{\mathrm{MA}}$ = main–auxiliary; $d_{\mathrm{LA}}$ = light–auxiliary. Ratio $r_v = d_{\mathrm{LA}}/d_{\mathrm{MA}}$.

Both main and light verb centroids remain substantially distant from the auxiliary centroid (mean $d_{\mathrm{MA}} = 0.637$; $d_{\mathrm{LA}} = 0.621$), while lying comparatively close to each other ($d_{\mathrm{ML}} = 0.177$). Four of the seven verbs yield ratios below 1.0: *uṭhā* (0.965), *diyā* (0.889), *gayā* (0.843), and *liyā* (0.929), indicating that their light-verb embeddings are displaced toward the auxiliary reference relative to their main-verb uses. The mean ratio is slightly below 1.0 ($r_v = 0.967$), but the direction varies across verbs: four ratios are below 1.0 and three are above it. Triangle visualizations showing all three models are provided in Appendix F.

## 6 Discussion

The results provide computational evidence consistent with three representational consequences derived from Butt's analysis of Urdu light verbs. First, main and light uses of the same verb are systematically differentiated in contextual embedding space. Second, despite this differentiation, same-lemma main and light centroids remain closer than mismatched main–light lemma pairs. Third, individual light verbs retain distinguishable contextual profiles after the target form is masked, including under a preceding-form-disjoint evaluation.

### 6.1 Main–Light Differentiation

Across all seven verbs and all three models, main and light uses occupy measurably different regions of embedding space. The centroid permutation tests show that this separation is unlikely under random label assignment, while the logistic probes establish that the distinction is linearly recoverable. These findings are compatible with Butt's claim that light uses make systematically different lexical-semantic and constructional contributions from their corresponding main uses.

The geometry of the distinction nevertheless varies across models. UrduBERT combines large centroid distances, higher silhouette scores, high probe accuracy, and comparatively strong KMeans alignment. DunbaaBERT, by contrast, achieves high probe F1 despite small centroid distances and near-zero mean ARI. This dissociation shows that representational differentiation need not take the form of two widely separated spherical clusters: a distinction may be linearly accessible while remaining geometrically compact or distributed across multiple sub-regions.

### 6.2 Lexical Relatedness Across Uses

For every model and verb, the matching main–light pair is closer than mismatched cross-lemma pairs, and each main centroid retrieves its corresponding light centroid at rank one. This pattern is consistent with continued lexical relatedness between main and light uses despite their contextual differentiation, as proposed by Butt and Lahiri (2013). The bootstrap intervals show that the cross-minus-same difference is stable under resampling.

The result does not directly establish a shared formal lexical entry. The same surface form is present in both matching conditions, and contextual models preserve token identity as well as contextual information. Accordingly, the proximity analysis should be read as a representational correlate compatible with lexical relatedness, not as proof of a particular lexical architecture.

### 6.3 Verb-Specific Structure Among Light Uses

The masked-target experiment provides a complementary test. When the target verb is removed, all models predict light-verb identity above both majority and uniform-chance baselines. This demonstrates that the contexts associated with individual light verbs are not interchangeable. They encode systematic differences in lexical selection, event structure, construction type, and distributional preference. This result is consistent with corpus-based evidence that individual Hindi light verbs impose distinct selectional constraints on their V1 complements (Vaidya et al., 2019).

UrduBERT's result is especially notable. Its masked accuracy is 0.866, and it retains 0.782 accuracy under the preceding-form-disjoint split. Thus, its performance is not explained solely by memorization of repeated immediate V1–V2 pairs. This generalization is consistent with Butt's view that light verbs retain verb-specific schematic contributions. DunbaaBERT also remains above baseline under the grouped split, while mBERT shows a larger decline, suggesting stronger dependence on recurring local lexical combinations.

These results do not isolate semantic content from all other contextual information. The classifier may exploit broader lexical, syntactic, genre, or constructional cues. The appropriate conclusion is therefore that individual light verbs have distinct and partially generalizable contextual profiles, a result compatible with verb-specific schematic event-structural content.

### 6.4 Graded Representational Separation

Representational separation varies across verbs, with *paṛā* showing the largest distances and several other verbs showing smaller values. This pattern may be compatible with graded semantic weakening, but embedding distance alone does not establish a scale of semantic distance. Independent human judgments or controlled semantic similarity experiments would be required to determine whether greater representational displacement corresponds to greater semantic reduction.

The verb-specific interpretations remain linguistically suggestive. For example, light *paṛā* frequently marks sudden or involuntary onset, whereas light *diyā* often preserves a causative or transfer-related schema. These observations motivate further semantic evaluation but are not independently validated by the present geometric measurements.

### 6.5 Auxiliary Reference Condition

The tense-auxiliary comparison indicates that light-verb centroids remain substantially distant from the auxiliary centroid across all models. For UrduBERT, four of the seven verbs yield relative auxiliary-proximity ratios below 1.0, indicating that their light-verb embeddings are displaced toward the auxiliary reference relative to their main-verb uses. This directional pattern is consistent with partial semantic reduction. However, both main and light centroids remain substantially more distant from the auxiliary than from each other, confirming that the main–light distinction is the dominant axis of variation. Because tense auxiliaries and light verbs differ in lexical identity and grammatical function, the analysis is treated as exploratory rather than as a direct measure of semantic distance.

## 7 Conclusion

This study investigated representational predictions derived from Butt's analysis of Urdu light verbs using 1,126 naturally occurring sentences and three contextual encoders. Main and light uses of the same seven verb forms are systematically differentiated in embedding space, although the magnitude and geometry of this distinction vary across verbs and models.

Two additional analyses strengthen the theoretical interpretation. Same-lemma main and light centroids are consistently closer than mismatched cross-lemma pairs across all models, a pattern compatible with continued lexical relatedness. Individual light verbs also remain recoverable from masked contexts, and UrduBERT retains high performance when immediate preceding surface forms are disjoint between training and test data. This indicates that the light-verb inventory preserves verb-specific and partially generalizable contextual structure rather than behaving as a homogeneous class of interchangeable markers.

Taken together, the findings provide computational evidence consistent with specific consequences of Butt's account: differentiation between main and light uses, continued lexical relatedness across those uses, and preservation of verb-specific schematic structure among light verbs. They do not directly prove the internal structure of lexical entries, establish a diachronic pathway, or provide a numerical measure of semantic distance.

## Limitations

The study is restricted to seven canonical Urdu light verbs, and its conclusions should not be generalized to the full light-verb inventory or to other languages. The data are naturally occurring rather than experimentally controlled, so lexical, syntactic, genre, and source-related cues may contribute to the observed embedding geometry.

The same-lemma proximity analysis retains the overt target form, meaning that shared orthographic and lexical identity may partly explain why matching pairs are closer than mismatched pairs. It therefore supports a prediction of representational relatedness but does not directly test a shared formal lexical entry. Similarly, the masked-target classifier demonstrates distinct contextual profiles but cannot isolate semantic information from all constructional or distributional cues.

The grouped masked evaluation uses automatically derived immediately preceding surface forms. Although it prevents exact surface-form overlap across folds, inflectional variants and complex verbal chains may require further manual normalization. Future work should repeat the analysis with manually verified V1 lemmas and construction types.

The tense-auxiliary comparison is exploratory and cannot establish a diachronic pathway or quantify semantic reduction because auxiliaries and light verbs differ in lexical identity, morphology, syntax, and grammatical function.

# A Ethical Consideration

This study uses publicly available Urdu text and involves no human participants or private data. All GPT-generated labels were independently reviewed by two Urdu linguistics experts, and only unanimously agreed instances were retained.

# B Computational Cost

UrduBERT was pre-trained on a single NVIDIA H100 GPU for approximately three days. The reported experiments required approximately 30 minutes while using roughly half of the GPU's available memory.

# C Use of AI-Assistant

Claude was used for coding, proofreading, and improve the text of this paper by correcting grammatical, spelling, and stylistic errors.

# D UrduBERT: Pre-training Details

## D.1 Pre-training Corpus

The UrduBERT pre-training corpus was compiled from multiple sources including the OSCAR 2019 corpus (Abadji et al., 2022), GitHub repositories, and Kaggle datasets. After collection, sentence-level deduplication was applied, yielding a final corpus of approximately 5.8 GB of Urdu text. The corpus was divided into five training files and one validation file of approximately 1.16 GB each, corresponding to a training-to-validation ratio of approximately 5:1.

## D.2 Tokenizer

A WordPiece tokenizer was trained on the pre-training corpus using the `BertWordPieceTokenizer` implementation from the HuggingFace Tokenizers library, with a vocabulary size of 64,000 tokens and a maximum sequence length of 512 tokens.

## D.3 Model Architecture

The UrduBERT model follows the standard BERT-base architecture (Devlin et al., 2019). The configuration is summarized in Table 8.

| **Hyperparameter** | **Value** |
|---|---|
| Hidden size | 768 |
| Number of layers | 12 |
| Attention heads | 12 |
| Intermediate size | 3,072 |
| Max position embeddings | 512 |
| Vocabulary size | 64,000 |
| Type vocabulary size | 2 |

Table 8: UrduBERT model configuration.

## D.4 Pre-training Procedure

The model was pre-trained using masked language modeling (MLM) with a masking probability of 0.15. Training was conducted using the HuggingFace Transformers `Trainer` API with the following hyperparameters:

| **Hyperparameter** | **Value** |
|---|---|
| Epochs | 20 |
| Per-device batch size | 24 |
| Gradient accumulation steps | 16 |
| Effective batch size | 384 |
| Learning rate | $1 \times 10^{-4}$ |
| Weight decay | 0.01 |
| Warmup steps | 10,000 |
| Precision | FP16 |
| Random seed | 42 |

Table 9: UrduBERT pre-training hyperparameters.

Training was performed on a single NVIDIA H100 GPU and required approximately 3–4 days to complete. The `get_last_checkpoint` utility was used to enable resumption from intermediate checkpoints, with checkpoints saved every 2,000 steps and evaluated every 4,000 steps.

## D.5 Downstream Evaluation

To validate the quality of the pre-trained model, UrduBERT was evaluated on two downstream tasks: Named Entity Recognition (NER) and Sentiment Analysis. In both tasks, UrduBERT outperformed multilingual BERT (mBERT), demonstrating that Urdu-specific pre-training yields representations better suited to Urdu language understanding. Detailed results will be released with the accompanying repository.

## E Main and Light-Verb Examples

Table 10 presents one representative main and light use for each target verb.

## F Visualizations

Figures 3–5 present per-verb triangle visualizations for all three models. Vertices = main (blue), light (red), AUXT (green); edge labels = cosine distance. Ratio $r_v$ above each subplot: green = $r_v < 1.0$, red = $r_v > 1.0$.

| Verb | Main usage | Light usage |
|---|---|---|
| *aaya* | رات کو اژدہا ہماری تلاش میں آیا<br>*raat ko azhdaha hamaari talaash mein aaya*<br>‘At night, the python came looking for us.’ | وہ اٹھا اور فلٹر سے صاف پانی لے آیا<br>*vo utha aur filter se saaf paani le aaya*<br>‘He got up and brought clean water from the filter.’ |
| *utha* | وہ فوراً اٹھا<br>*vo foran utha*<br>‘He got up immediately.’ | دل جھوم اٹھا<br>*dil jhoom utha*<br>‘The heart suddenly rejoiced.’ |
| *baitha* | اس کے بعد یہ صرف وہیں بیٹھا<br>*us ke baad yeh sirf waheen baitha*<br>‘After that, he simply sat there.’ | مَیں اپنے ہوش کھو بیٹھا<br>*main apne hosh kho baitha*<br>‘I ended up losing my senses.’ |
| *para* | ہندی میں بو کو گند کہتے ہیں جس سے اس کا نام گندھک پڑا<br>*Hindi mein boo ko gand kehte hain, jis se us ka naam gandhak para*<br>‘In Hindi, smell is called *gand*, from which it received the name *gandhak*.’ | عیسیٰ رو پڑا<br>*Isa ro para*<br>‘Isa burst into tears.’ |
| *diya* | کس نے اسے نوبل دیا<br>*kis ne use Nobel diya*<br>‘Who gave him the Nobel Prize?’ | آپ نے پھینک دیا<br>*aap ne phenk diya*<br>‘You threw it away.’ |
| *gaya* | آج میں بازار گیا<br>*aaj main bazaar gaya*<br>‘Today I went to the market.’ | ٹھیک ہو گیا<br>*theek ho gaya*<br>‘It became all right.’ |
| *liya* | للی نے کافی کا پہلا گھونٹ لیا<br>*Lily ne coffee ka pehla ghoont liya*<br>‘Lily took the first sip of coffee.’ | پولیس نے مقدمہ درج کر لیا<br>*police ne muqadma darj kar liya*<br>‘The police registered the case.’ |

Table 10: Representative main and light uses of the seven target verbs. Each example is presented in Urdu script, followed by transliteration and English translation.

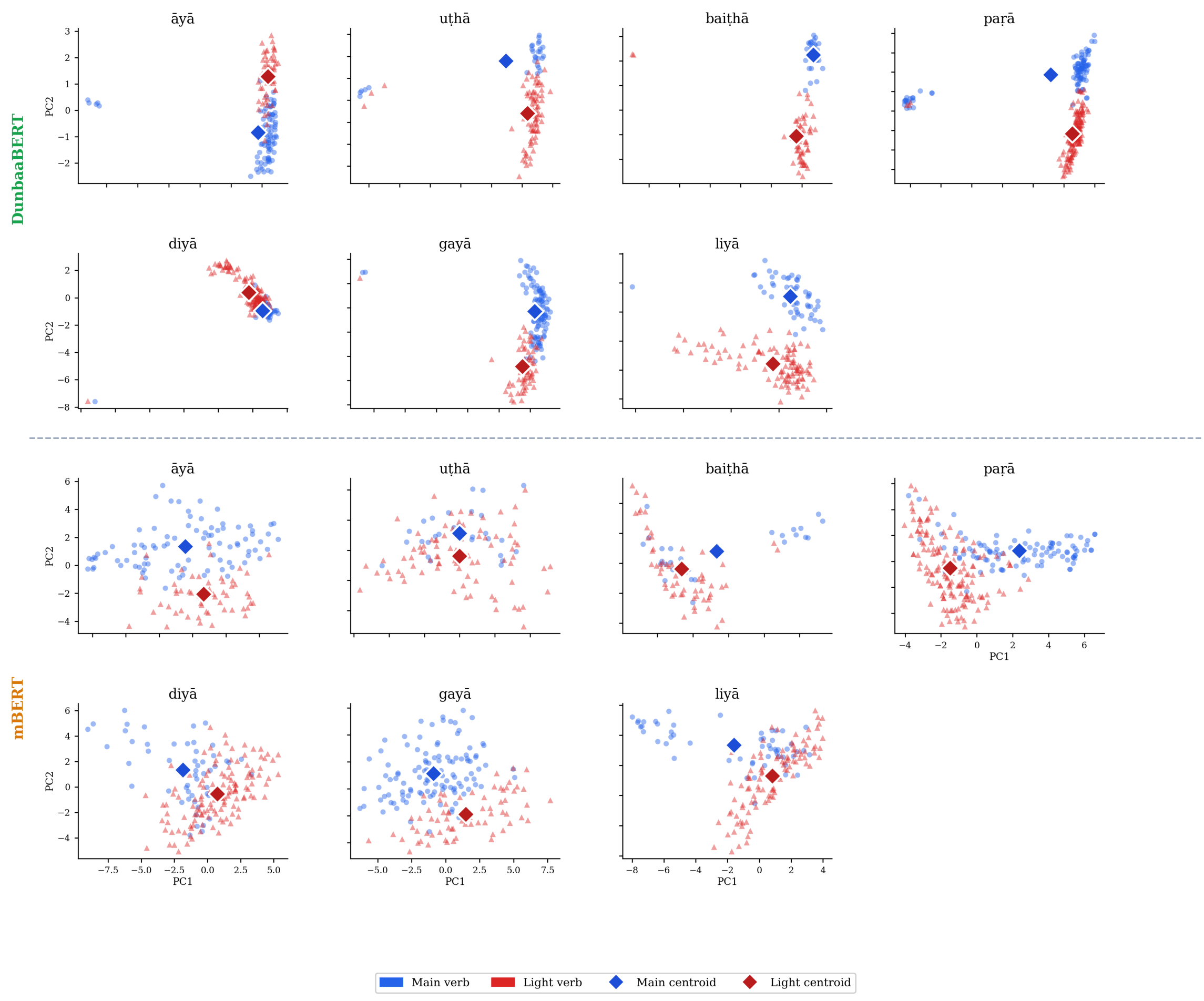

Figure 2: Per-verb PCA projections for DunbaaBERT and mBERT.

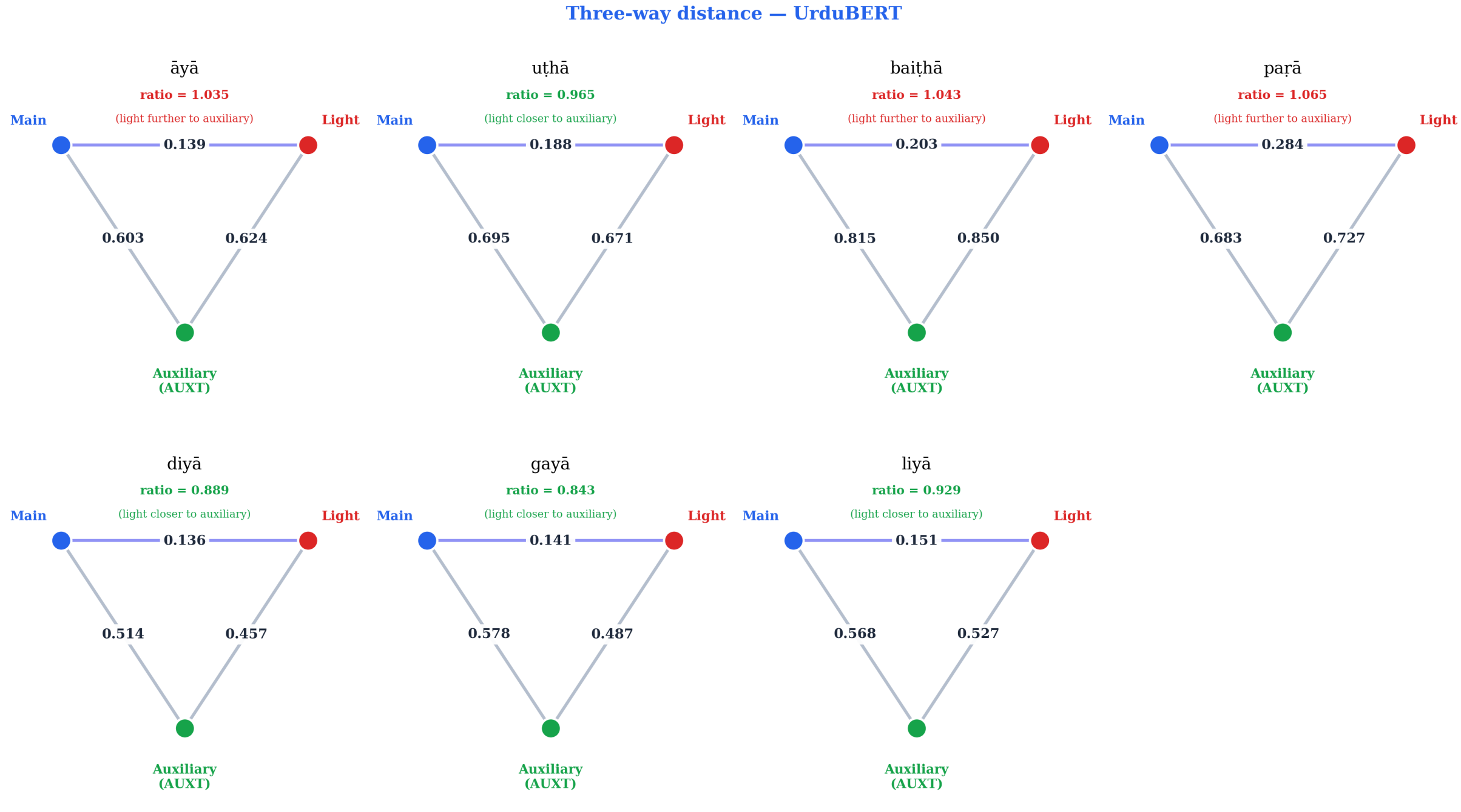

Figure 3: Three-way distance — UrduBERT.

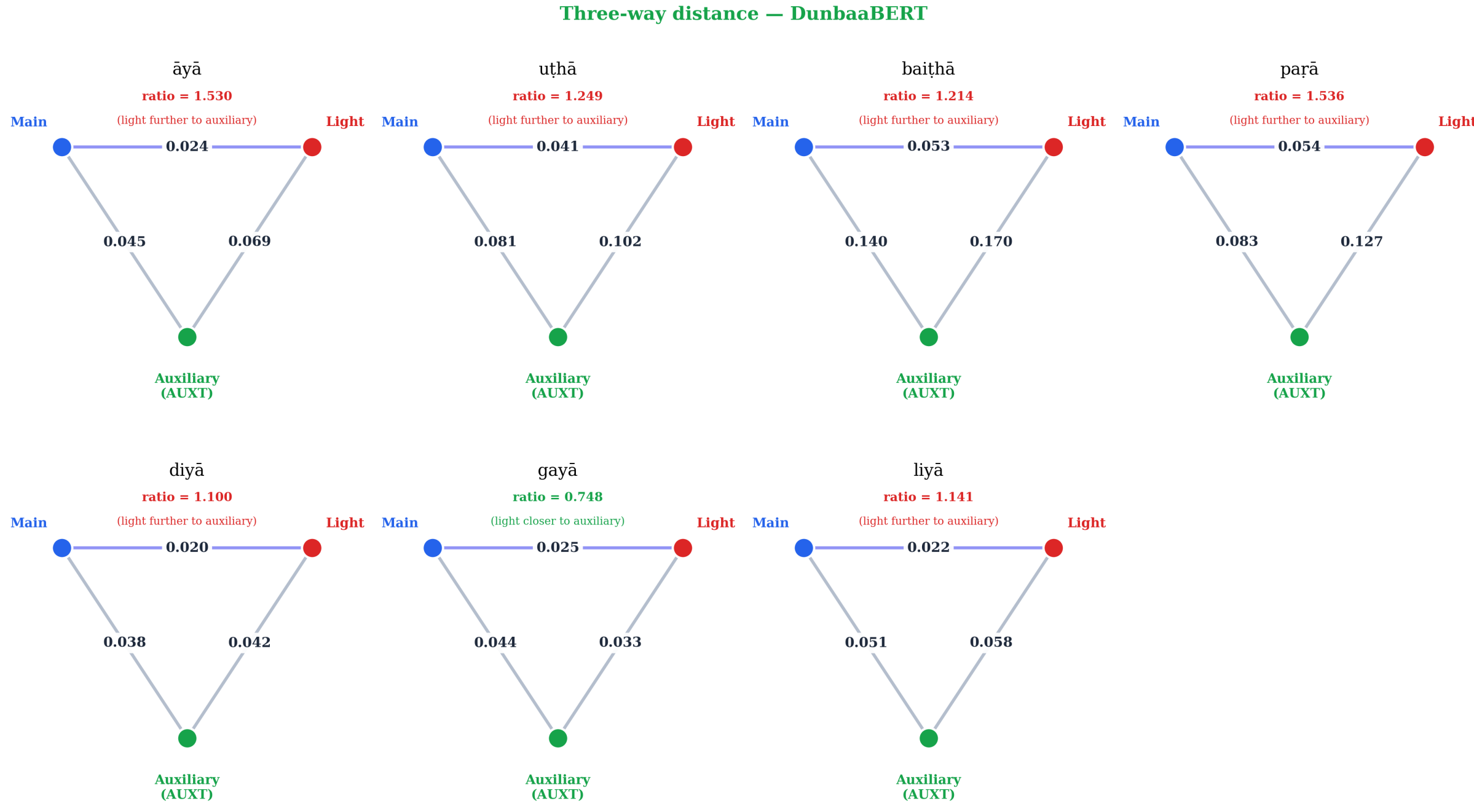

Figure 4: Three-way distance — DunbaaBERT.

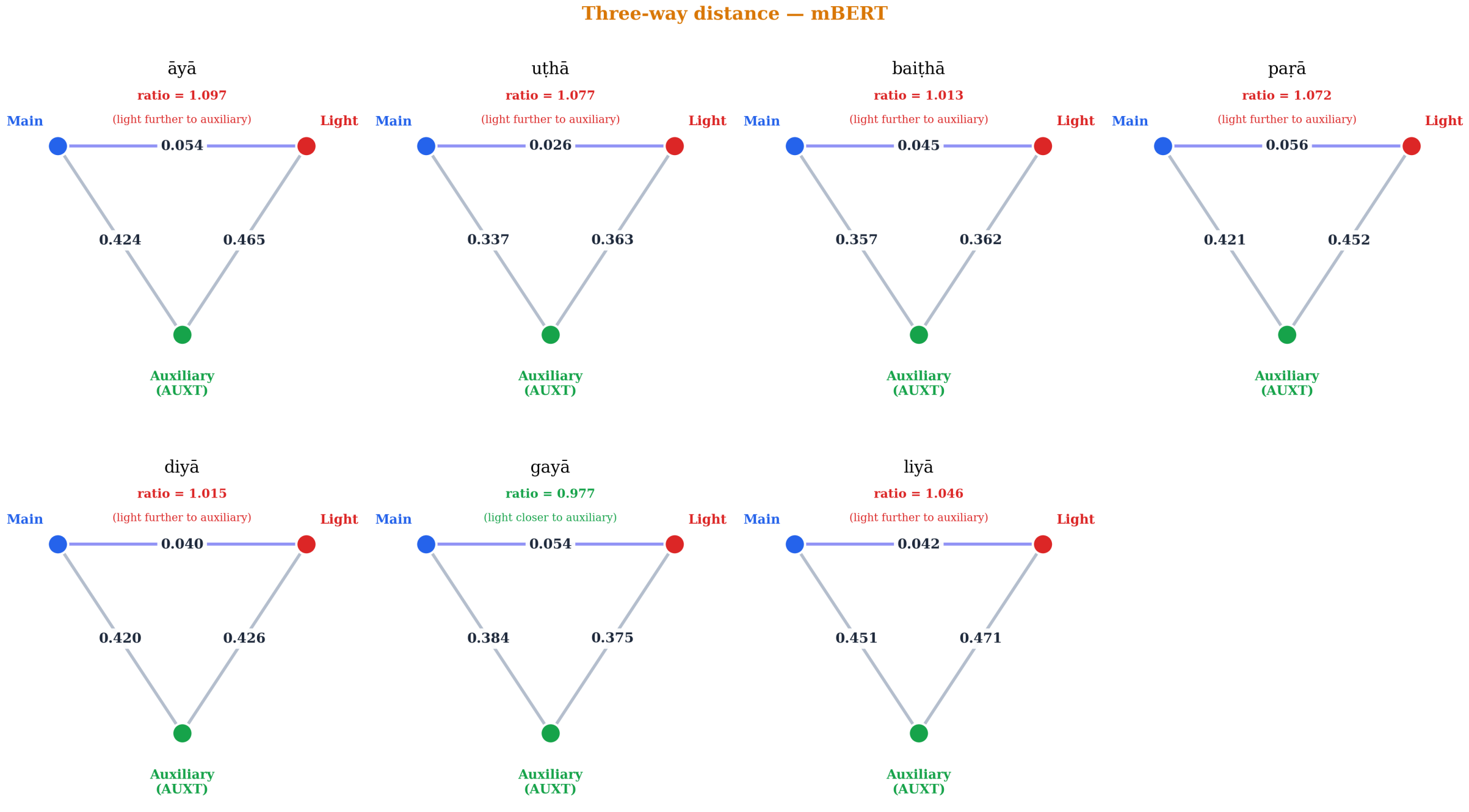

Figure 5: Three-way distance — mBERT.